# Multi-step prediction of chlorophyll concentration based on Adaptive Graph-Temporal Convolutional Network with Series Decomposition

**Ying Chen, Xiao Li, Hongbo Zhang, Wenyang Song and Chongxuan Xv**

School of Electrical Engineering, Yanshan University, Qinhuangdao, Hebei, China

E-mail: chenying@ysu.edu.cn

## Abstract

Chlorophyll concentration can well reflect the nutritional status and algal blooms of water bodies, and is an important indicator for evaluating water quality. The prediction of chlorophyll concentration change trend is of great significance to environmental protection and aquaculture. However, there is a complex and indistinguishable nonlinear relationship between many factors affecting chlorophyll concentration. In order to effectively mine the nonlinear features contained in the data. This paper proposes a time-series decomposition adaptive graph-time convolutional network ( AGTCNSD ) prediction model. Firstly, the original sequence is decomposed into trend component and periodic component by moving average method. Secondly, based on the graph convolutional neural network, the water quality parameter data is modeled, and a parameter embedding matrix is defined. The idea of matrix decomposition is used to assign weight parameters to each node. The adaptive graph convolution learns the relationship between different water quality parameters, updates the state information of each parameter, and improves the learning ability of the update relationship between nodes. Finally, time dependence is captured by time convolution to achieve multi-step prediction of chlorophyll concentration. The validity of the model is verified by the water quality data of the coastal city Beihai. The results show that the prediction effect of this method is better than other methods. It can be used as a scientific resource for environmental management decision-making.

## 1. Introduction

The water environment is one of the fundamental components of the environment and a crucial setting for the survival and advancement of human civilisation. At the same time, humans have significantly harmed and disrupted the aquatic ecosystem [1]. Excess nutrients in the water column are extremely likely to cause algal bloom outbreaks in shallow nearshore areas, bays, and estuaries with inadequate water exchange capacities [2]. However, hypernutrition in coastal waters has been documented worldwide [3,4]. One of the most severe issues in coastal waters nowadays is hypernutrition. Chlorophyll concentration is regarded as the most direct representation for determining the danger of overnutrition by a number of widely used composite trophic status indices [5,6]. Predicting chlorophyll concentrations has become a crucial component of early warning systems used to stop or reduce algal blooms [7]. Through the sensor and network, we can monitor water quality, and the current water quality data prediction is mostly short-term. This will certainly cause a lag in water quality control and be prone to sudden water quality pollution accidents, seriously affecting the timeliness of production decisions. Therefore, multi-step prediction of chlorophyll concentration through modeling of water quality parameters is of great significance. Chlorophyll concentration prediction is a typical multivariate time series prediction task, i.e., constructing a mapping model from historical data of multiple water quality parameters to future chlorophyll concentration data [8].

ARIMA and SVR are examples of common modern statistical prediction algorithms[9]. ARIMA models are used to learn the linear association between features. However, the data must be stable for the ARIMA model [10] to work. In the case of linear prediction, the SVR prediction model can achieve good results, but for multi-parameter complex issues, SVR additionally has to conduct kernel function determination and quadratic programming, and the prediction results are not sufficient [11].

Machine learning techniques can easily handle nonlinear data and complex hydrological and environmental processes, overcoming the limitations of traditional models [12,13], and have been widely used in water quality prediction [14-16]. For instance, utilizing different input variables, RF and ANN have been used to estimate water levels in rivers. Yajima et al. used RF models to forecast trends in chlorophyll concentrations in water [17]. Kim et al. created ANN models to assess altering river systems based on chlorophyll concentrations [18].

Driven by computer development and research enthusiasm, deep learning has developed rapidly and has been more widely used in marine environments [19], such as ship navigation [20], wave [21], sea surface temperature [22], and some studies on predicting ocean chlorophyll concentration [23]. Recurrent neural networks (RNNs) and convolutional neural networks (CNNs) are the two main branches of deep learning [24,25]. RNNs have an embedded feedback, recursive structure, which allows them to preserve information from previous moments and use the previous information to predict the current information [26]. However, RNNs have some drawbacks in terms of gradient transfer and cannot solve the long-term dependency problem. The LSTM and GRU models retain the advantages of RNN and mitigate the shortcomings of gradient transfer to some extent [27,28]. Cen et al. developed a single-input single-output LSTM prediction model for chlorophyll-a concentrations in the East China Sea, which displayed good results while illustrating the shortcomings of long-term prediction accuracy degradation due to not accounting for the effects of other parameters [29]. Jiang et al. evaluated the effects of multi-source urban data on water quality prediction in sewage networks, finding that GRU has greater prediction ability and a faster learning curve [30]. Convolutional neural networks (CNNs) are mostly utilized in image processing. A one-dimensional convolutional neural network (1D-CNN) is a form of convolutional neural network that is better suited for sequential data processing [31]. BAI et al. later suggested a temporal convolutional network model that may be employed as a general convolutional structure model for sequential data processing [32]. Kang et al. successfully predicted ammonia nitrogen content in water using temporal convolutional networks [33]. However, using a single algorithm to predict water quality has the drawback of not fully utilizing the distinctive patterns of data. To fully take advantage of the underlying feature patterns in the data, hybrid models can combine the benefits of several techniques [34]. Hybrid model research has become a hotspot for solving difficult time series challenges. Barzegar et al. employed a hybrid CNN-LSTM deep learning model to predict short-term water quality variables and demonstrated that the hybrid model can capture more information [35]. Graph neural networks (GNNs) are frequently employed to address the limitations of temporal models since they are efficient at mining spatial characteristics of data. However, in some sectors, it might be challenging to predefine graph structures in a reasonable manner, which makes it challenging for GNNs to perform well.

In view of the above problems, this paper proposes a time series decomposition adaptive graph-time convolution network ( AGTCNSD ) prediction model. Through sequence decomposition, we can understand the attributes of the original data in more detail and conduct comprehensive data analysis. At the same time, the structure of the graph neural network is improved, and a parameterized embedding matrix is defined to build an adaptive learning graph structure. The graph structure of the adaptive graph convolution is learnable, which not only solves the problem of graph structure that is difficult to determine, but also uses the learning relevance of the graph network. At the same time, the idea of matrix decomposition is used to assign different learnable weight parameters to each node to improve the learning ability of the update relationship between nodes. Then, the temporal convolutional network captures the long-term dependence of the data. Each of these three modules is designed to extract a specific feature pattern. Taking the data of Beihai coastal waters as an example, the proposed prediction method is verified by comparing several baseline methods.

## 2. Relative work

### 2.1 Sequence decomposition method

Sequence decomposition is one of the commonly used methods in time series analysis. Time series has complex components and is generally considered to be the superposition or coupling of the following types of changes : long-term trends ( Trend, T ), seasonal changes ( Seasonal, S ), periodic changes ( Periodic, P ) and irregular changes ( Irregular, I ). The commonly used sequence decomposition methods include STL decomposition, empirical mode decomposition ( EMD ) and moving average decomposition. STL is a time series decomposition method based on robust local weighted regression as a smoothing method. Empirical mode decomposition is a method for processing non-stationary signals. The key of this method is empirical mode decomposition. It can decompose complex signals into a finite number of intrinsic mode functions ( IMF ). The decomposed IMF components contain local characteristic signals of different time scales of the original signal. The moving average method uses the sliding window to gradually move along the time direction of the time series, and calculates the average number of data within a fixed window size in turn to reflect the long-term trend of the sequence.

## 2.2 Graph convolutional neural networks

One of the biggest advances in deep learning in recent years is the expansion of deep learning into the field of graph, that is, graph deep learning. The graph provides a general representation of the data. Data from many systems from different domains can be explicitly represented as graphs, such as social networks, traffic networks, protein-protein interaction networks, knowledge graphs, and brain networks. At the same time, many other types of data can also be converted into graphs. A large number of real-world problems can be handled as a set of computational tasks on a graph. Such as node classification problems such as inferring node attributes, detecting abnormal nodes, and identifying disease-related genes. Node connection prediction problems such as recommendation, drug side effect prediction, drug-targeted interaction recognition, and knowledge map completion. The nodes of the graph are essentially connected, which means that the nodes are not independent and distributed differently. Therefore, the graph data structure essentially contains many data relationship expression capabilities that traditional learning techniques do not have.

## 2.3 Time convolution neural network

Convolutional neural network is a kind of neural network with convolution calculation and deep structure. Similar to other neural network algorithms, the input features of convolutional neural networks need to be standardized due to the use of gradient descent algorithms for learning. Specifically, before inputting the learning data into the convolutional neural network, it is necessary to normalize the input data in the channel or time / frequency dimension. If the input data is pixels, the original pixel values distributed in [0,255] can also be normalized to the [0,1] interval. The standardization of input features is conducive to improving the learning efficiency and performance of convolutional neural networks. The hidden layer of convolutional neural network includes three common structures : convolutional layer, pooling layer and fully connected layer. In some more modern algorithms, there may be complex structures such as residual blocks. Convolution layer and pooling layer are unique to convolutional neural networks. The convolution kernel in the convolution layer contains the weight coefficient, while the pooling layer does not contain the weight coefficient, so the pooling layer is not considered to be an independent layer.

Convolutional neural network was first applied in the field of image. With the development of convolution, convolution can also be used in sequence data, and the key is the use of one-dimensional convolution. The simplest one-dimensional convolution calculation principle is consistent with the ordinary convolution, but it only limits the width and movement direction of the convolution kernel. The width of the convolution kernel is the same as the input dimension, and the convolution operation is carried out along the time dimension of the time series.

## 3. Methods

### 3.1 Data sources

The experimental data in this study are from the coastal waters of Beihai, Guangxi Province, from January 1, 2020, to December 31, 2022. Beihai is a picturesque seaside city, and the region has a thriving tourism and mariculture business. However, anthropogenic production activities tend to change the quality of water bodies. The investigation of water quality in this area is important for both the preservation of the environment and the growth of the local economy.

The data acquisition platform adopts EM700 ( Fig.1(a) ) of Xiandao Instrument and Equipment Co., Ltd. The components of EM700 buoy include GPS, wireless MODEM antenna, radar reflector, solar panel, data acquisition electronic warehouse, battery, buoy body, heavy vertical stability connecting rod, anchor and anchor rope, YSI-6600 ( multi-parameter water quality analyzer, Fig.1(b) ). In order to meet the needs of water quality monitoring, the detection buoy is equipped with high-precision and high-performance YSI series sensors, which can not only measure conventional parameters, but also simultaneously measure parameters including optical dissolved oxygen, blue-green algae, turbidity, chlorophyll and rhodamine. YSI 6600V2 multi-parameter water quality monitor can effectively measure optical dissolved oxygen, blue-green algae, turbidity, chlorophyll and other factors.

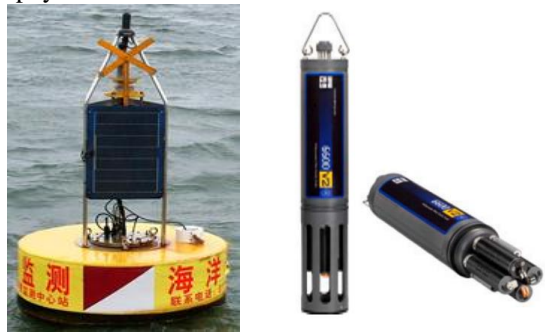

(a) EM700          (b) YSI 6600V2

**Fig.1** Equipment physical diagram in sensor network

The ocean monitoring buoy fitted with the YSI water quality parameter sensor collects data. The sensors record measurements every 30 minutes, and the measured parameters include temperature, SpCond/mS, Cond/mS (conductivity), Sal (salinity), TDS/ppt (total dissolved solids), DO/% (dissolved oxygen saturation), DO/ppm (mg/L) (dissolved oxygen), pH, pH/mV, Turb/NTU (turbidity), Chl

/ppb (ug/L) (chlorophyll), Chl/RFU (relative fluorescence units), PE/uL, PE/RFU (phycohemoglobin). The distribution and change of water quality parameters obtained by the sensors are depicted in Fig.1. There are some outliers and missing values in the data, which will affect the upcoming data analysis and modeling work.

### 3.2 Data pre-processing

Data pre-processing is a crucial step in the preparation of work involving data analysis and modeling. The failure of transmission lines and data collection equipment, among other mechanical factors, frequently results in abnormal or missing data records. A preliminary analysis of the experimental data revealed some jump spots and interruptions that substantially exceeded the acceptable range and did not follow the variation law of ocean parameters. Anomalies in data and missing data can be handled in one of two ways: directly deleting them, or replacing the erroneous data with valid information[36]. To keep the data's temporal consistency, the second method is applied to time series data. The pauta criterion is a quick and accurate method for identifying outliers. The data is a gross error and is regarded as an outlier if the absolute gap between it and the mean is larger than three times the standard deviation. The missing data can be swiftly and logically filled using segmented linear interpolation. As a result, the data was completely filled using the segmented linear interpolation approach after the pauta criterion was applied to identify all outliers and replace them with null values. After some basic pre-processing, the data visualization is shown in Fig.3.

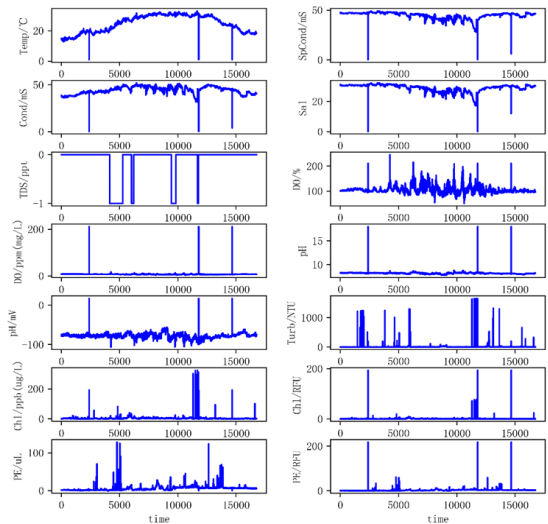

**Fig.2** Visualization of raw data

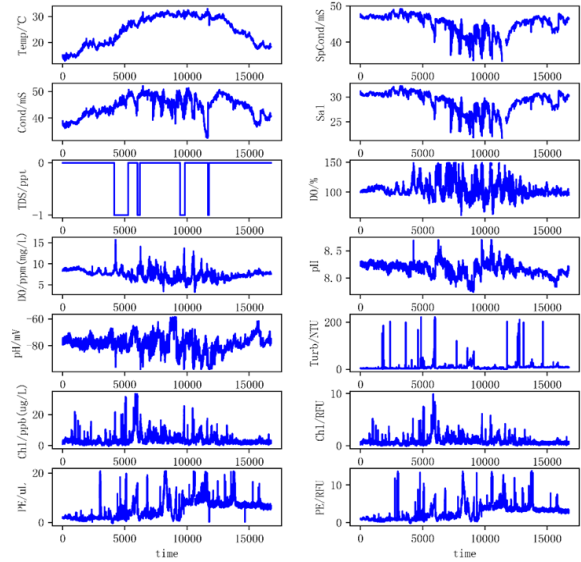

**Fig.3** Visualization of pre-processed data

In the water environment, water quality parameters have a limited range of continuous changes in a short period of time, so the data at a certain moment can represent the state of water quality in that period of time. The sensor recording frequency is relatively dense, so the original data is sampled, which has the advantage of reducing the training difficulty of the model and doubling the prediction effectiveness by predicting the same number of steps.

The water quality parameters illustrated in Fig.3 differ by a huge order of magnitude, and direct input into the training model will reduce the impact of data of a lower order of magnitude. Data normalization can convert dimensioned data to dimensionless data and transform data with varying magnitudes to the same magnitude. To normalize the data, min-max normalization is employed, and the rationale is to scale the data between 0 and 1.

### 3.3 Feature Selection

Feature selection has the function of reducing redundant, irrelevant, noisy and uninformative data, which can reduce storage space and reduce time complexity. At the same time, dimension reduction helps to alleviate the problem of overfitting. The two intertwined problems of overfitting and slow training can be alleviated by feature reduction.

The comparability of multivariate sequences after data standardization is enhanced. Next, the importance of various water quality parameters to chlorophyll prediction is briefly analyzed by Pearson coefficient method :

$$r = \frac{\sum_{i=1}^{n}(x_i - \overline{x})(y_i - \overline{y})}{\sqrt{\sum_{i=1}^{n}(x_i - \overline{x})^2}\sqrt{\sum_{i=1}^{n}(y_i - \overline{y})^2}} \quad \ldots\ldots(1)$$

When the Pearson coefficient is in the range of [ -1,1 ], the greater the absolute value, the higher the correlation. 0-0.2 is extremely weak or irrelevant, 0.2-0.4 is weak correlation, 0.4-0.6 is moderate correlation, 0.6-0.8 is strong correlation, 0.8-1 is extremely strong correlation. The parameters with the absolute value of the correlation coefficient greater than 0.2 are selected for training the model. The order of importance of water quality parameters for chlorophyll concentration prediction is shown in Fig.4.

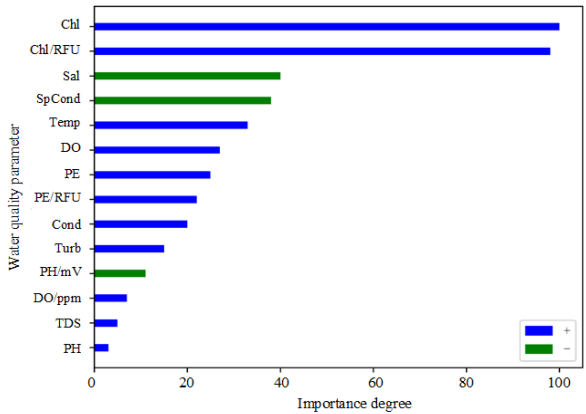

**Fig.4** Importance ranking of water quality parameters

Among them, '+' represents that the water quality parameter has a positive correlation with chlorophyll concentration, and '-' represents that the water quality parameter has a negative correlation with chlorophyll concentration. If the correlation coefficients of the two water quality parameters to the chlorophyll concentration are the same, the effect of the prediction is the same, and one of them can be regarded as redundant data. Therefore, the selected model inputs are Chl, Sal, SpCond, Temp, Do, PE and Cond.

### 3.4 Building AGTCNSD

#### 3.4.1 Model Framework.

The structural flow of AGTCNSD is depicted in Fig.5. AGTCNSD is made up of three modules: series decomposition module(Fig.3(a)), graph convolution module (Fig.3(b)), and temporal convolution module (Fig.3(c)).

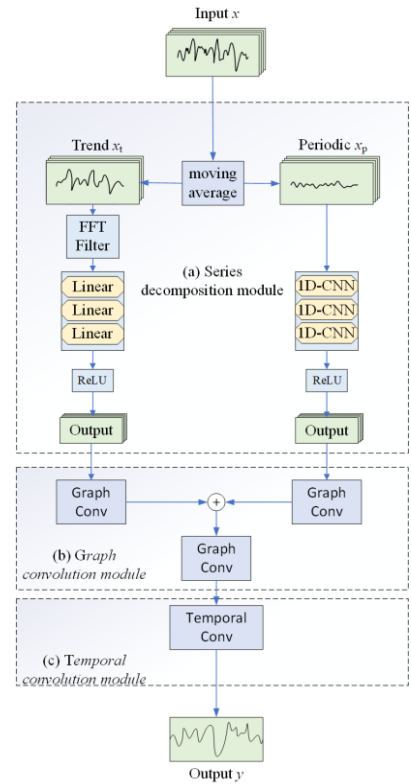

**Fig.5** Overall structural flow of AGTCNSD

After series decomposition, the data are first divided into trend and period components. The trend component represents the overall trend of the series, and the trend component's global information is initially integrated after linear transformation; the period component contains the series' period variation and random noise, and the random noise is filtered using fast Fourier transform to get a purer period component, and the period variation is initially sensed by 1D-CNN. Next, the graph convolution module receives the data after the initial processing by the series decomposition module, adaptively learns the correlation between parameters, and updates the state of each parameter according to the correlation. Finally, the temporal dependence of the data is extracted by the temporal convolutional network, and the prediction results are output.

#### 3.4.2 Series decomposition module.

Sequence decomposition is one of the commonly used methods in time series analysis [ 36,37 ]. The commonly used sequence decomposition methods include STL decomposition, empirical mode decomposition ( EMD ) and moving average decomposition. STL is a time series decomposition method based on robust local weighted regression as a smoothing method. Empirical mode decomposition is a method for processing non-stationary signals. The key of this method is empirical mode decomposition. It can decompose complex signals into a finite number of intrinsic mode functions ( IMF ). The decomposed IMF components contain local characteristic signals of different time scales of the original signal. The moving average method uses the sliding window to gradually move along the time direction of the time series, and calculates the average number of data within a fixed window size in turn to reflect the long-term trend of the sequence.

Generally, sequence decomposition will be implemented in data preprocessing, and the sequence will be decomposed into multiple or a series of sequence components. If the dimension of the original sequence is large, the amount of data will increase rapidly after decomposition. It is necessary to classify and manage different types of components, which is subjective and suitable for data expansion of small data sets. Since the moving average decomposition method is simple and efficient, it is easy to modularize it. Therefore, this paper constructs a moving average decomposition module as an internal block of deep learning. The algorithm is simple and efficient, and easy to embed into the model. It can decompose the original time series into trend and periodic components in order to use information from different angles. The internal structure of the sequence decomposition module is shown in Fig. 6.

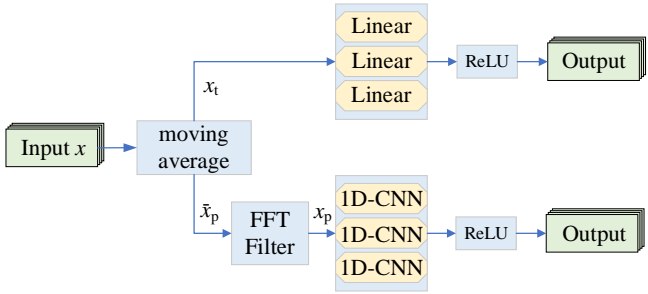

**Fig.6** Internal structure of the series decomposition module

The trend component $x_t$ of the series is first calculated by moving average of the input series, and then the trend component is subtracted from the original series to obtain the periodic component $\tilde{x}_p$ containing random noise:

$$x_t = \text{movingaverage}(x) \qquad (2)$$

$$\tilde{x}_p = x - x_t \qquad (3)$$

where $\text{movingaverage}$ is the moving average operation. The periodic component is smaller in order of magnitude compared to the trend component, and the random error mixture is more influential in the periodic component, so it is important to filter out the random error in the periodic component and obtain a purer periodic component. Fast Fourier transform is one of the important tools in the field of signals and is widely used in the analysis of time series data [37, 38]. The fast Fourier transform is used to filter the periodic components. The principle is as follows: First, the original sequence $\hat{x}_p(t)$ is converted to frequency domain signal $\hat{X}(\omega)$ by fast Fourier transform. Then calculate its power spectrum $P(\omega)$ to determine the contribution of each frequency component, and select the top k frequency components $\hat{X}(\omega)$ with the largest contribution as the main body of the periodic component. Then the Fourier inverse transform is performed to obtain the pure periodic component $x_p(t)$. The formulas are:

$$\hat{X}(\omega) = \int_{-\infty}^{\infty} \hat{x}_p(t) e^{-j\omega t} dt \qquad (4)$$

$$P(\omega) = \lim_{T \to \infty} \frac{\left| \hat{X}(\omega) \right|^2}{2\pi T} \qquad (5)$$

$$X(\omega) = \hat{X}(\omega)[Top_k(P(\omega))] \qquad (6)$$

$$x_p(t) = \frac{1}{2\pi} \int_{-\infty}^{\infty} X(\omega) e^{j\omega t} d\omega \qquad (7)$$

The effect of moving average decomposition and noise filtering is shown in Fig.7.

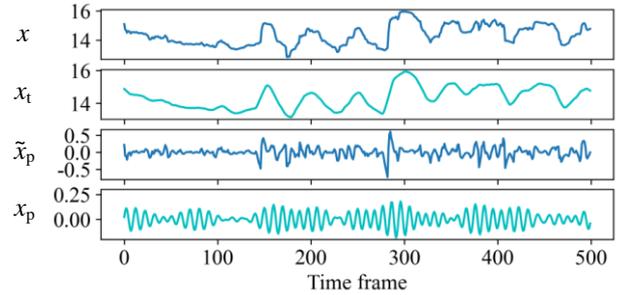

**Fig.7** Effect of sequence decomposition and noise filtering

For the trend component, the linear neural network is used for initial processing because the linear layer can learn the global features of the data. For periodic components, the 1D cellular neural network is used for preliminary processing because the convolution structure is more sensitive to local feature changes. The convolution kernel sizes of the three 1D cellular neural networks are set to 3,5 and 7, respectively, so that the model can learn a wide range of local features. Through sequence decomposition, we can analyze data features in detail from the perspectives of trend and cycle. Targeted modular feature extraction is conducive to reducing the difficulty of feature extraction by only one subject module.

### 3.4.3 Graph convolution module.
Convolutional graph network ( GCN ) is a model for processing unstructured data. Because GCN has obvious advantages in processing data correlation, it plays an irreplaceable role in user product recommendation, molecular structure prediction and traffic [41-43]. In this paper, the graph convolutional neural network is used to model the water quality parameter data, and each water quality parameter is regarded as a node of the graph. The graph convolutional neural network is shown in Fig.8. However, in some fields, graph convolution does not work better if the graph structure of the data is unknown, that is, the adjacency matrix representing the graph structure cannot be determined. Therefore, we define a parameterized embedding matrix and an adaptive learning graph structure. In the adaptive graph structure, the node updates its own state according to its neighbor nodes, but the weight used to update the state of each node is shared, and the learning efficiency is low. Therefore, the idea of matrix decomposition is used to assign different learnable weight parameters to each node to improve the learning ability of the update relationship between nodes.

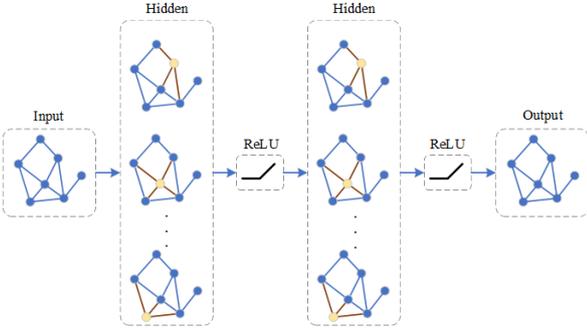

**Fig.8** Basic structure of adaptive graph convolution

The graph convolution operation can be well approximated by a first-order Chebyshev polynomial expansion and extended to high-dimensional GCNs [42], as follows:

$$Z = (I_N + D^{-\frac{1}{2}}AD^{-\frac{1}{2}})X\Theta + b \qquad (8)$$

where $A$ is the adjacency matrix of the graph, $D$ is the degree matrix, $X$ and $Z$ are the inputs and outputs of the graph convolution, and $\Theta$ and $b$ are the learnable weights and biases. In this model, both the adjacency matrix and the degree matrix are defined and fixed. To enable the graph convolution to learn the graph structure of the data adaptively, the parameterized matrix $E_A$ is predefined to replace the part of the adjacency matrix and the degree matrix multiplication, and the spatial correlation between each pair of nodes is inferred by multiplying $E_A$ and $E_A^T$, then the adaptive adjacency matrix $A_{ad}$ can be expressed as:

$$A_{ad} = D^{-\frac{1}{2}}AD^{-\frac{1}{2}} = softmax(ReLU(E_A \cdot E_A^T)) \qquad (9)$$

At the same time, in order to reduce the number of learnable parameters and better exploit the features of mutual influence between graph nodes, the matrix decomposition idea is used to decompose $\Theta$ and $b$, assigning different parameters to each node, and the comprehensive influence between parameters is more explanatory. Then the adaptive GCN can be expressed as:

$$Z = (I_N + A_{ad})XE_\varsigma W_\varsigma + E_\varsigma b_\varsigma \qquad (10)$$

where $E_\varsigma$, $W_\varsigma$ and $b_\varsigma$ are the decomposition terms of the weights and biases.

In general, the definition of parameter space is randomly generated. Although this can avoid the influence of the initial distribution of parameters on the model, in certain cases, even if the random range is controlled within a small range, it is easy to cause problems such as gradient disappearance and gradient explosion, which makes the model unable to train. Therefore, when defining the parameter space, we define the parameter space of the normal distribution, and the deep model can be well trained. Therefore, adaptive graph convolution has greater flexibility to update the relevant states of nodes in the actual network topology.

### 3.4.4 Temporal convolution module.

The temporal convolutional network mainly consists of causal convolution and dilated convolution, as shown in Fig.9. The causal convolution ensures that future information will not leak into the past. The dilation convolution exponentially expands the sensory field layer by layer. Causal convolution and dilation convolution give temporal convolution a superior ability to handle time series. Compared to recurrent neural networks, TCNs have a longer memory capacity thanks to their larger receptive fields. For a one-dimensional series input $x \in R^n$ and a convolution kernel $f : \{0, \cdots, k-1\} \rightarrow R$, the dilated convolution operation $F$ on the series element $s$ is defined as:

$$F(s) = (x *_d f)(s) = \sum_{i=0}^{k-1} f(i) \cdot x_{s-d\cdot i} \qquad (11)$$

where $d$ is the dilation factor, $k$ is the filter size, and $s\text{-}d\cdot i$ denotes the past direction.

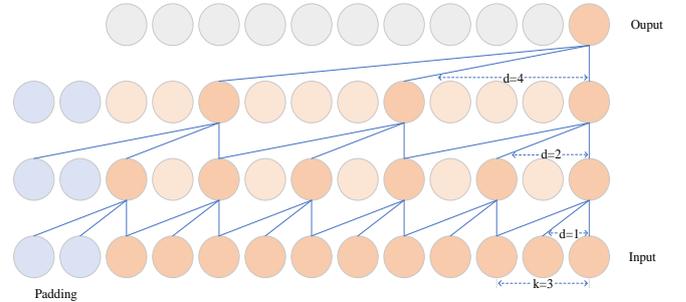

**Fig.9** Temporal convolutional network

Convolutional networks learn data features through convolutional kernels, which have the advantages of parameter sharing and a small number of parameters[45]. Therefore, convolutional structures tend to increase data dimensionality in the intermediate layers in order to learn richer data features. In this paper, a dimensionality reduction linear layer is added before each convolutional layer to fuse and compress the data dimensions. While retaining the information richness, the data in the model is made lighter, and the performance of the model is improved.

## 4. Experiment and Discussions

### 4.1 Experimental environment and setup

The hardware environment for this experiment is a computer environment with Intel(R) Core(TM) i7-11800H @ 2.30GHz and an NVIDIA GeForce RTX 3060 Laptop GPU card, and the software environment is NVIDIA-cuda 11.3, python 3.8.13, and the deep learning framework PyTorch-GPU 1.11.0.

**Table.1** Model parameter settings

| module | model parameter | setting |
|---|---|---|
| / | training batch size | 128 |
| | epochs | 300 |
| | learning rate | 0.001 |
| | optimization algorithm | Adam |
| series decomposition module | moving average window size | 12 |
| | number of periodic frequency selection | 15 |
| | 1D-CNN kernel size | [3,5,7] |
| graph convolution module | number of layers | 2 |
| | parameterized matrix size | (14,7) |
| | decomposition matrix size | (14,7),(7,14) |
| temporal convolution module | number of layers | 4 |
| | kernel size | 3 |
| | dilation factor | [1,2,4,8] |

Experimental data construction. For the multi-step prediction problem of chlorophyll concentration, the input length of the model is fixed, and the output is the predicted value of chlorophyll concentration corresponding to the future multi-step time point. Considering the actual problems, the data acquisition frequency of the ocean buoy monitoring system is 30min / time, and the water quality parameters change weakly in such a short time. Therefore, the pre-processed data is sampled before constructing the data set. The sampled data can be regarded as the data collected by the ocean buoy monitoring system at a frequency of 1h / time. The amount of data is halved and can well reflect the changes of water quality parameters. The data of each time step represents an increase in the duration from 30 min to 1h. Therefore, the same number of prediction steps will double the future time represented before, so that the multi-step prediction results can reflect a longer chlorophyll change trend. In addition, we set the sliding window size of four target sequences, which are 12,24,48 and 72, respectively. It represents the prediction range of half a day, one day, two days and three days in the future. The model parameters are shown in Table 1.

### 4.2 Baseline model and evaluation criteria

This paper selected two widely used deep learning models as the baseline models. They are the LSTM model and the TCN model. LSTM, as a representative model of a recurrent neural network in deep learning, has been applied and achieved good results in many fields. The TCN model is a representative model of deep learning for convolutional structured networks in sequential data tasks with extra-long temporal memory capability and outperforms the LSTM model in certain sequential tasks. As two important branches of deep learning, the LSTM and TCN models are chosen to be representative of the baseline models.

In this paper, root mean square error (RMSE), mean absolute error (MAE), and mean absolute percentage error (MAPE) are chosen as evaluation criteria to measure the performance of each method. MAE indicates the mean of the absolute error between the predicted and true values, which can effectively reflect the prediction error. RMSE indicates the dispersion between the predicted and true values, which can measure the presence of outliers in the predicted values. MAPE can describe the accuracy of the prediction, and the smaller the three indicators, the better. They are calculated as follows:

$$MAE = \frac{1}{n}\sum_{i=1}^{n}\left|(y_i - \hat{y}_i\right| \qquad (12)$$

$$RMSE = \sqrt{\frac{1}{n}\sum_{i=1}^{n}(y_i - \hat{y}_i)^2} \qquad (13)$$

$$MAPE = \frac{1}{n}\sum_{i=1}^{n}\left|\frac{(y_i - \hat{y}_i)}{y_i}\right| \qquad (14)$$

### 4.3 Ablation experiments

In order to verify the effect of the sequence decomposition module, the period enhancement module and the graph convolution module, we designed the ablation experiment with 24-step prediction.

(1) TCN model without structural changes : Model 1 ;

(2) TCN model with structural changes : Model 2 ;

(3) Remove the sequence decomposition module in the AGTCNSD model : Model 3 ;

(4) Remove the time convolution module in the AGTCNSD model : model 4 ;

(5) Remove the graph convolution module in the AGTCNSD model : model 5 ;

(6) Complete AGTCNSD model : Model 6 ;

**Table.2** Results of ablation experiment

| model | MAE | RMSE | MAPE |
|---|---|---|---|
| model1 | 0.4642 | 0.5823 | 0.2165 |
| model 2 | 0.4500 | 0.5795 | 0.2108 |
| model 3 | 0.3901 | 0.4979 | 0.1809 |
| model 4 | 0.3428 | 0.4415 | 0.1703 |
| model 5 | 0.4381 | 0.5277 | 0.2082 |
| model 6 | 0.3360 | 0.4209 | 0.1526 |

The experimental results are shown in Table.2. The experimental results show that both the series decomposition module and the adaptive graph convolution module play a useful role. The focus of these two modules is different. When added to the time convolution prediction model, the prediction effect can be significantly improved. When the two modules are used simultaneously, the prediction results are further improved. This shows that the hybrid network can take advantage of each component, and components with different focuses will not conflict. The effectiveness of these modules is proved.

## 4.4 Results and Discussion

The AGTCNSD, LSTM, and TCN models were built in the experiments. The three models were trained and tested using water quality data. The experimental results are in Table.3.

Overall, the AGTCNSD model outperformed the LSTM and TCN models in terms of prediction. The AGTCNSD model does not show a great advantage over the TCN model in the 12-step prediction, indicating that the detailed information extracted by AGTCNSD is not very useful in the shorter-term prediction. When the number of prediction steps increases, the advantage of AGTCNSD is clearly shown. It indicates that the model is able to learn the trend of the data effectively, ensuring that the trend does not produce huge deviations in the output prediction, and that the learning of the periodic components produces some intervention in the overall trend prediction, ensuring the existence of reasonable fluctuations in the details of the predicted data. The large number of parameters that can be learned by the graph convolutional network enables better learning of how each parameter is affected by the other parameters. This detailed information effectively suppresses the problem of rapid degradation of the model's prediction performance due to increasing prediction step sizes.

**Table.3**. Performance results of AGTCNSD compared with the baseline models

| Models | 12 steps | | | 24 steps | | |
|---|---|---|---|---|---|---|
| | MAE | RMSE | MAPE | MAE | RMSE | MAPE |
| LSTM[26] | 0.4918 | 0.5590 | 0.2376 | 0.8068 | 0.8867 | 0.3888 |
| TCN[31] | **0.2845** | **0.3578** | 0.1346 | 0.8332 | 0.9077 | 0.4031 |
| AGTCNSD | 0.2867 | 0.3583 | **0.1285** | **0.3360** | **0.4209** | **0.1526** |

| Models | 48 steps | | | 72 steps | | |
|---|---|---|---|---|---|---|
| | MAE | RMSE | MAPE | MAE | RMSE | MAPE |
| LSTM | 0.7776 | 0.7617 | 0.3851 | 0.8659 | 0.9431 | 0.4321 |
| TCN | 0.8503 | 0.9609 | 0.3580 | 0.7510 | 0.7840 | 0.2767 |
| AGTCNSD | **0.4004** | **0.4938** | **0.1858** | **0.5388** | **0.6528** | **0.2479** |

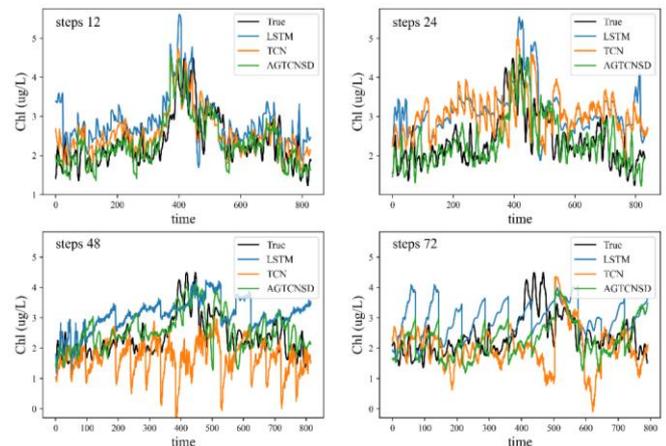

**Fig.10** Comparison of multi-step prediction results of different models

To facilitate observation, we visualized the predictions at corresponding prediction steps in the test set and compared all the test results with the true values, as shown in Fig.10. In the 12-step prediction, all three models can predict the overall trend of the true value better, and our model deviates the least. Starting from 24-step-ahead prediction, the LSTM and TCN models have shown significant distortion in prediction, while AGTCNSD still maintains good prediction performance. In 48-step prediction, LSTM and TCN have completely lost their prediction ability and deviated from the trend change of the real data, and AGTCNSD also showed some deviation in a few cases. In the 72-step prediction, AGTCNSD also appears to fail at some times, and around 500 time frames, the true value appears to be higher in the region, and AGTCNSD can also give an approximate simulation, indicating that the prediction ability has not been completely lost. However, as the prediction step increases, the prediction becomes more and more difficult, becomes less effective, and is no longer sufficient to support environmental management decisions.

The evaluation performance indexes of the experimental test results indicate that the AGTCNSD model has excellent results and good stability. Combined with the visual analysis of the prediction results and the increase in prediction step length, the indexes for evaluating the performance of the model do not drop rapidly with the 72-step ahead prediction, but the prediction effect has not reached the level of prediction warning. Overall, our model has obvious advantages over the two representative models for both short-term and long-term forecasting.

## 5. Conclusion

This article studies the prediction technology of chlorophyll concentration in water bodies. Considering the complex nonlinear relationship between multiple parameters in water and the lag in taking measures to address water quality deterioration, an AGTCNSD model for multi-step prediction of chlorophyll concentration was proposed. The model evaluation experiment was conducted using water quality data along the North Sea coast, and the experimental results and analysis were presented. The AGTCNSD model utilizes a sequence decomposition module to decompose the input sequence into trend components and periodic components, allowing for detailed analysis of the sequence from two aspects; The period enhancement module makes the period components observable, eliminates noise interference, and enables the model to better grasp the fluctuations of the sequence; Graph convolution learns the relationship between different water quality parameters on the graph structure, and updates the state of its own nodes using the influence of neighboring nodes, which can effectively extract effective information from multiple aspects; Finally, changing the internal parameters of time convolution makes the model more suitable for multi-step prediction. The experimental results show that the AGTCNSD model can learn the trend and periodic fluctuations of the sequence, which normalizes the prediction range to a considerable range without generating significant deviations. Graph convolutional networks can learn and integrate data on graph structures, and their self-adaptability eliminates the influence of subjective factors and enhances the interpretability of the model. Due to the increase in output steps, time convolution also needs to provide more dimensional pattern information for prediction, which is beneficial for the output layer to map features to the target space. The AGTCNSD model performs well in multi-step prediction tasks. As the prediction length increases, the model can also maintain a certain level of predictive ability and provide reasonable prediction results within a certain prediction duration. However, as a long-term prediction model, it still cannot reach the level of early warning. Overall, it provides scientific reference for water environment management and effectively solves the lag in water environment protection and decision-making.

## 6. Future work

It is of great practical significance to apply deep learning technology to the research of water quality prediction. The single-point accuracy of single-step prediction is very impressive, but the information provided by single-step prediction is relatively limited. Therefore, multi-step prediction is more practical. However, although multi-step prediction can describe the general trend of the sequence in the future, the accuracy of single point is relatively poor. Even the deep learning model with many parameters can not perfectly solve the problem of multi-step prediction. The feature representation ability of deep learning is unquestionable, and the tasks of deep learning will be more complex in the future, such as multiple multi-step prediction tasks. With the continuous development of deep learning and more and more complex task requirements, some small but indispensable elements also need to keep pace with the times, such as activation functions and training strategies.

## Acknowledgements

This work was supported by the National Natural Science Foundation of China under Grant 62275228 and the Key Research and Development Project of Hebei Province under Grant 19273901D and 20373301D.